\documentclass[11pt,a4paper]{article}
\usepackage{acl-hlt2011}
\usepackage{times}
\usepackage{latexsym}

\usepackage{url}  %Required
\usepackage{graphicx}  %Required
\usepackage{amsmath}
\DeclareMathOperator*{\argmax}{arg\,max}

\usepackage{adjustbox}
\usepackage{multirow, array}
\usepackage{makecell}
\usepackage{tabularx}
\usepackage{lipsum}

\newcommand\blfootnote[1]{%
  \begingroup
  \renewcommand\thefootnote{}\footnote{#1}%
  \addtocounter{footnote}{-1}%
  \endgroup
}

\begin{document}
\title{Multi-Instance Learning for End-to-End Knowledge Base Question Answering}
\author{Mengxi Wei\footnote{}\\
  Beijing University of \\ Posts and Telecommunications \\
  {\tt wmx@bupt.edu.cn} \\\And
  Yifan He \hspace{5mm} Qiong Zhang \hspace{5mm} Luo Si\\
  Alibaba Group \\
  {\tt y.he@alibaba-inc.com} \\}

%\author{Mengxi Wei\footnote{Work done at Alibaba Group} \hspace{5mm} Yifan He \hspace{5mm} Qiong Zhang \hspace{5mm} Si Luo\\
%Alibaba Group\\
%Beijing University of Post and Telecommunication\\
%\texttt{wmx@bupt.edu.cn, y.he@alibaba-inc.com}
%%2275 East Bayshore Road, Suite 160\\
%%Palo Alto, California 94303\\
%}
\maketitle
\begin{abstract}
% End-to-end approaches to knowledge base question answering (KBQA) have the advantage that they
% can be trained directly on QA pairs, but user generated answers are of varied quality. Noisy
% answers may adversely affect training.
End-to-end training has been a popular approach for knowledge base question answering (KBQA). However, real world applications often contain answers of varied quality for users' questions. It is not appropriate to treat all available answers of a user
question equally.

This paper proposes a novel approach based on multiple instance learning to address the problem of noisy answers by exploring consensus among answers to the same question in training end-to-end KBQA models. In particular, the QA
pairs are organized into bags with dynamic instance selection and different options of instance weighting. Curriculum learning is utilized to select instance bags during training. On the public CQA dataset, the new method significantly improves both entity accuracy and the Rouge-L score over a state-of-the-art end-to-end KBQA baseline.
% We also significantly outperform baselines
% in accuracy and naturalness when the data is composed of both single- and multiple-answer
% questions.

\end{abstract}
\section{Introduction}
%The development of
%scalable and efficiently adaptable KBQA systems is an important research problem,
%as various applications in different domains,
%from customer service to semantic search, would benefit from KBQA sub-modules.
% In many use cases such as customer service, the task involves understanding the
% question, retrieving facts from the KB, and generating natural language answers.
Knowledge Base\blfootnote{* Work done during Mengxi Wei's internship at Alibaba Group.} Question Answering (KBQA) aims to build systems that automatically respond to
natural language questions with
natural answers using information in knowledge bases. It has been an active research topic for a long time~\cite{bobrow1977gus}. Traditionally,
systems are built as a pipeline of components, including NLU, KB retrieval, reasoning,
and answer generation. The components can either base on rules or statistical models~\cite{bobrow1977gus,grishman1979qa,zelle1996geoquery,zettlemoyer2005ccg,tur2016multi}.

\begin{figure}[ht]
\centering
\includegraphics[width=0.58\columnwidth]{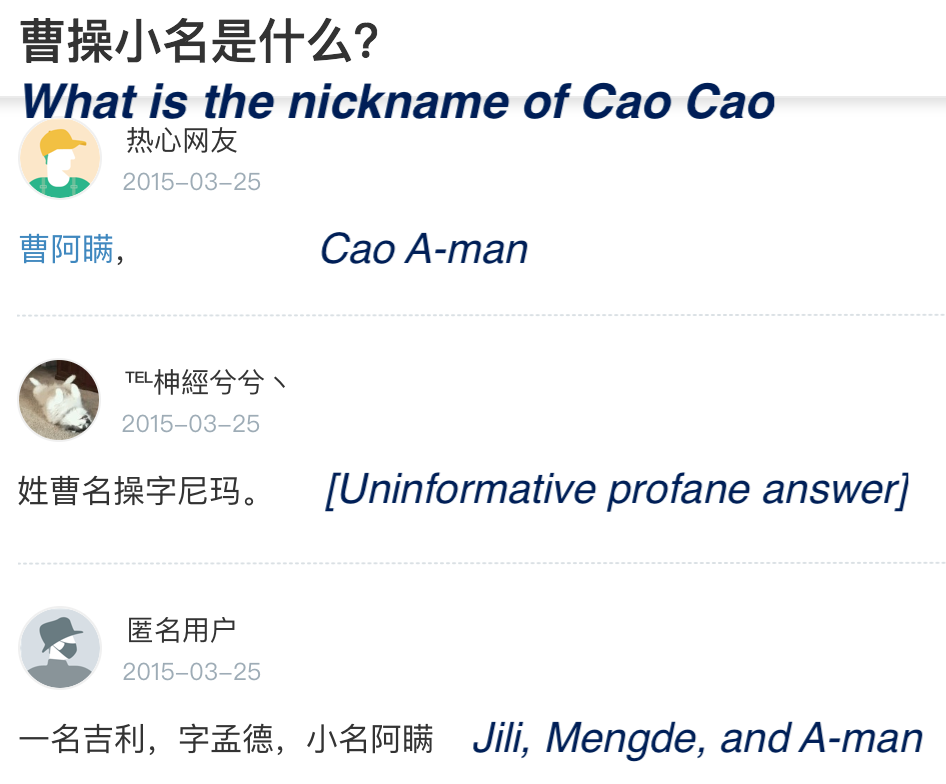}
\caption{Screenshot of a community QA website.}\label{f:zhidao}
\end{figure}

One limitation of the pipeline approach is that it requires much resource to scale up or adapt
to different domains. Moving the pipeline to new domains often means preparing new sets of rules
and/or annotating new data to train different statistical models, both of which are difficult and time-consuming.
% Moreover, pipeline systems are also prone to error propagation, when errors in
% upstream components hurt the performance of downstream components.

End-to-end KBQA approaches~\cite{yin2016genqa,he2017coreqa}
have been attracting interests in recent years, because
they can be trained directly with QA pairs collected from the web.
These are easier to acquire than e.g. NLU frame annotations for pipeline KBQA systems.
User-generated QA data are usually abundant and have facilitated training of
several successful KBQA models~\cite{yin2016genqa,he2017coreqa}.

However, the quality of user-generated question-answer pairs is not guaranteed. Figure
\ref{f:zhidao} is a screenshot from a community QA website.  We observe : 1) \textbf{Irrelevant answers}, e.g.
the second answer is a vulgar joke that does not contain any information that we
can find in a reasonable KB; and 2) \textbf{Inconsistent answers}. In Figure~\ref{f:zhidao}, the third answer covers more complete
information in the KB than the first. It is desirable to filter out
noisy responses and to promote high quality answers.

In this paper, we attempt to address these challenges by organizing data differently from
previous work. Instead of training
on QA pairs (as most end-to-end KBQA systems do to the best of our knowledge), we organize answers to the same
question into bags and train with the multi-instance learning principle. This formulation
allows us to select and weight answers according to consensus among
the answers in the same bag.

\begin{figure*}[h!]
\centering
\includegraphics[width=0.9\textwidth]{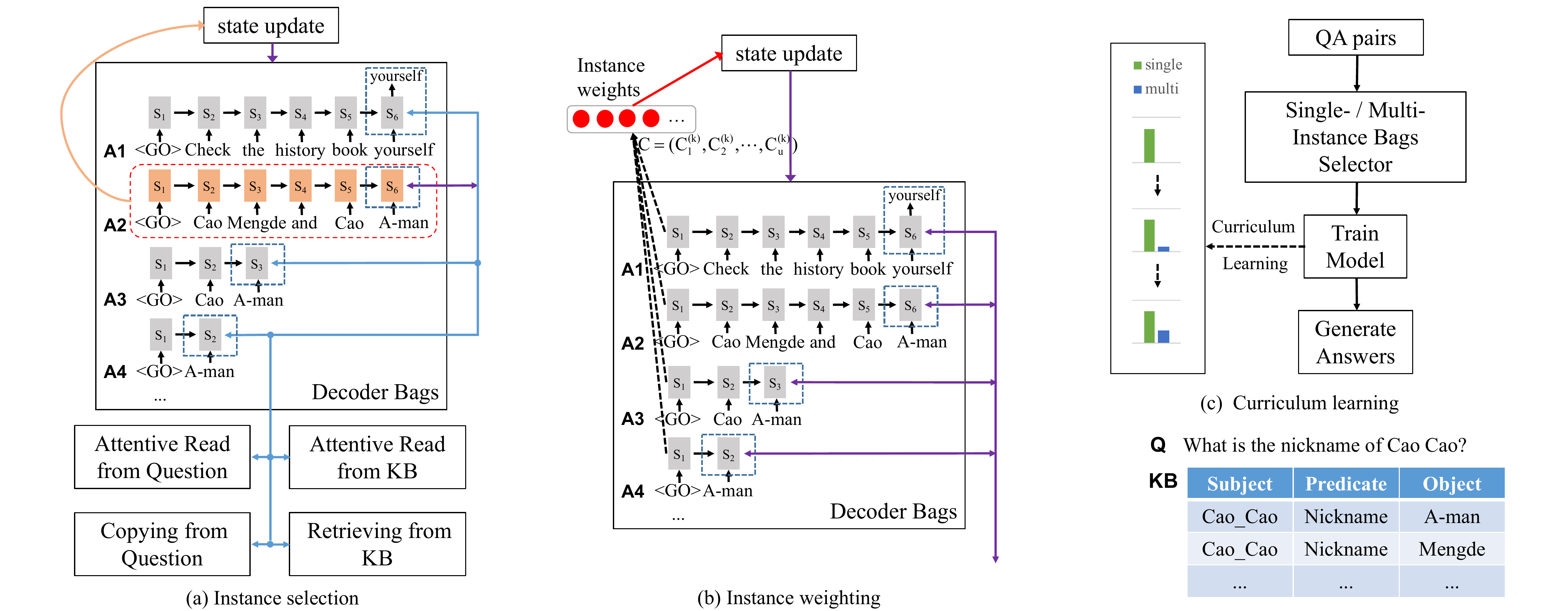}
\caption{Proposed multi-instance KBQA approaches.}\label{f:sys}
\end{figure*}
\section{Related Work}
A number of recent work~\cite{yin2016genqa,fu2018heteromem,lin2018dsqa,madotto18MemSeq,liu2018curriculum} has explored KBQA
using neural networks. GenQA~\cite{yin2016genqa} is among the first in this
strand of research that retrieves facts from the KB
and generates answers from these facts.
Our model follows the approach of CoreQA~\cite{he2017coreqa}, which improves GenQA
by allowing multiple attentive reads from both the KB and the question.

A related, but different question answering approach
is based on Machine Reading Comprehension (MRC: ~\cite{wang2018multi,yu2018qanet}). Recent MRC research utilizes
relation graphs among entities to integrate evidences in
text~\cite{song2018mrc}.
Our work is different from MRC in that the knowledge graph in KBQA
is manually curated, while MRC extracts the answer from free text.
% This line of research is further extended to incorporate more knowledge sources
% and generate more natural answers. For example, ~\cite{fu2018heteromem} incorporate
% knowledge from different sources and use cumulative attention to avoid repetition.
% ~\cite{liu2018curriculum} apply the curriculum learning principle to first train
% on simple QA pairs and gradually add more complex pairs, in order to generate more
% natural and informative answers. These work improves end-to-end KBQA
% in different ways, but they do not try to model the difference in the quality
% of user-generated answers.
% There are systems that try to solve problems related to KBQA: e.g. Mem2Seq~\cite{madotto18MemSeq}
% incorporates KBs into task-oriented dialogs using a memory module, but focus on multi-turn dialogs;
% DS-QA~\cite{lin2018dsqa} use paragraph selectors to filter out noise in machine
% reading-based open domain QA, which is different from our KB-based setting.

Multi-instance learning~\cite{dietterich1997mil} is a variant of supervised learning where
inputs are bags of instances.
One successful application in NLP is distant supervision of relation extractors by
only learning from some of the instances~\cite{riedel2010mil,zeng2015pcnmil},
or assigning different weights to the instances under mechanisms
such as selective attention~\cite{lin2016selective}. Our task is a generation problem, where
classification techniques developed for relation extraction cannot be applied directly.

We use curriculum learning to schedule training instances in our method. Curriculum
learning~\cite{bengio2009curriculum} is a paradigm that schedules simple
training examples during early epochs of training, and gradually adds hard examples
to the process. It has found successful applications in multi-task NLP~\cite{mccann2018decanlp} and
KBQA~\cite{liu2018curriculum}.

% \begin{equation}
% \begin{array}{rcl}
% \forall \textbf{z}: \textbf{z}^T \nabla ^2L_\textbf{w} \textbf{z}  & = & \textbf{z}^T (y\sigma (1-\sigma) \textbf{x} \textbf{x}^T) \textbf{z}\\
% & = & y\sigma (1-\sigma) \textbf{z}^T \textbf{x} \textbf{x}^T \textbf{z} \\
% & = & y\sigma (1-\sigma) (\textbf{x}^T \textbf{z})^2 \geq 0 \\
% \end{array}
% \end{equation}

\section{End-to-end KBQA}

The KBQA task takes a question $\mathbf{x}$ and a KB as input and
outputs a response $\mathbf{y}$ based on the information in the KB.

Our baseline KBQA model follows the CoreQA~\cite{he2017coreqa} approach.
The model encodes the question as
$\mathbf{M}_{q}$ with a bi-directional LSTM and
the KB as $\mathbf{M}_{kb}$ by concatenating the embeddings of the subject-predicate-object triples. The model decodes with an LSTM decoder
that generates from the output vocabulary and copy / retrieve from
$\mathbf{M}_{q}$ and $\mathbf{M}_{kb}$, in similar fashion to the pointer-generator
~\cite{see17ptrgen}.

\subsection{Question Encoder}

The question encoder encodes a question $\mathbf{x}$ into a vector encoding and builds
a short term memory $\mathbf{M}_q$ that covers every word in the question.

The question $\mathbf{x}$ is a word sequence $[x_1, \cdots , x_L]$ of length $L$. For a
word $x_t$, we use a bi-directional LSTM encoder to obtain its forward state
$\overrightarrow{\mathbf{h}}_t$ and its backward state $\overleftarrow{\mathbf{h}}_{L-t+1}$.
The short term memory of the sentence $\mathbf{M}_q$ is the concatenated forward and
backward states of the words, i.e. $\mathbf{M}_q=\{\mathbf{h}_t\}$, where $\mathbf{h}_t =
[\overrightarrow{\mathbf{h}}_t, \overleftarrow{\mathbf{h}}_{L-t+1}]$. The question
is encoded by the concatenation of the last states in both directions, i.e.
$\mathbf{q}=[\overrightarrow{\mathbf{h}}_t, \overleftarrow{\mathbf{h}_1}]$.

\subsection{KB Encoder}

The KB encoder encodes a KB (set of relation triples) into a short term memory
$\mathbf{M}_{kb}$.

The KB consists of relation triples of type \texttt{<s, p, o>}, in which
\texttt{s} (subject) is the entity, $p$ (predicate) is the name of the relation,
and $o$ (object)
is the value of the relation. For example, \texttt{<Cao Cao, nickname, A-man>}
is a fact about the entity \texttt{Cao Cao}, stating that his nickname is \texttt{A-man}.

Denoting the embeddings of \texttt{s}, \texttt{p}, and \texttt{o} as $\mathbf{e}_s$,
$\mathbf{e}_p$, and $\mathbf{e}_o$ respectively, we represent a fact with the concatenation
of these three embeddings: $\mathbf{f}=[\mathbf{e}_s, \mathbf{e}_p, \mathbf{e}_o]$.
We consider fact representations as the short term memory of the KB, i.e $\mathbf{M}_{kb}=
\{\mathbf{f}_1, \cdots , \mathbf{f}_{N}\}$, where $N$ is the number of facts.

\subsection{Decoder}

The decoder is an LSTM~\cite{hochreiter97lstm} network that generates answers from the question
encoding while attending to the
short term question
memory $\mathbf{M}_q$ and the short term KB memory $\mathbf{M}_{kb}$. The output at time
$t$ is a mixture of three modes: \textit{prediction} (pr), \textit{copy} (cp), and
\textit{retrieve} (re), as in Eq. (\ref{eq:coreqa}).

\begin{equation}\label{eq:coreqa}
\begin{array}{rcl}
p(y_t| \mathbf{s}_t, y_{t-1}, \mathbf{M}_q, \mathbf{M}_{kb}) & = & p_{pr} \cdot p_{m}(pr | \mathbf{s}_t, y_{t-1}) + \\
       &   & p_{cp} \cdot p_{m}(cp | \mathbf{s}_t, y_{t-1}) + \\
       &   & p_{re} \cdot p_{m}(re | \mathbf{s}_t, y_{t-1})   \\
\end{array}
\end{equation}

\noindent
where $p_{pr}$, $p_{cp}$, and $p_{re}$ are prediction mode, copy mode, and retrieval mode
output distributions respectively, and $p_{m}$ is the mode selector implemented as a 2 layer NN with
softmax activation.

The state of the RNN $\mathbf{s}_t$ is
updated with the previous state $\mathbf{s}_{t-1}$, the previously generated word,
and the attention context $\mathbf{c}_t$ based on attentive reads of the question
and the KB.

\subsubsection{Prediction mode} The prediction mode generates new words from the vocabulary,
considering attentive reads from the question and the KB memory, as in Eq.
(\ref{eq:coreqa-pr}).

\begin{equation}\label{eq:coreqa-pr}
p_{pr}(y_t=y_j| \cdot) = DNN_1(\mathbf{s}_t, \mathbf{c}_{qt}, \mathbf{c}_{kbt})
\end{equation}

\noindent
where $\mathbf{s}_t$ is the current state of the decoder, $\mathbf{c}_{qt}$ is the attentive
read from $\mathbf{M}_{q}$, and $\mathbf{c}_{kbt}$ is the attentive reading from
$\mathbf{M}_{kb}$.

\subsubsection{Copy mode} The copy mode measures the probability of copying word $x_j$ from
the question, as in Eq. (\ref{eq:coreqa-cp}).

\begin{equation}\label{eq:coreqa-cp}
p_{cp}(y_t=x_j| \cdot) = DNN_2(\mathbf{s}_t, \mathbf{h}, \mathbf{hist}_{q})
\end{equation}

\noindent
where $\mathbf{h}$ is the attentive read of the question memory $\mathbf{M}_{q}$ and $\mathbf{hist}_q$
is the accumulated attention history on the question~\cite{tu2016coverage}.

\subsubsection{Retrieval mode} The retrieval mode measures the probability of retrieving the
predicate value $o_j$ from the KB, as in Eq. (\ref{eq:coreqa-re}).

\begin{equation}\label{eq:coreqa-re}
p_{re}(y_t=o_j| \cdot) = DNN_3(\mathbf{s}_t, \mathbf{f}, \mathbf{hist}_{kb})
\end{equation}

\noindent
where $\mathbf{f}$ is the attentive read of
$\mathbf{M}_{kb}$ and $\mathbf{hist}_{kb}$ is the accumulated attention history on the
KB triples.

In the three modes, $DNN_1$, $DNN_2$, and $DNN_3$ are 2 layer MLPs with softmax activation.

CoreQA is trained on the negative log likelihood loss. Given questions $\mathbf{x}_1 \cdots
\mathbf{x}_m$ and answers $\mathbf{y}_1 \cdots \mathbf{y}_m$, the negative log
likelihood loss sums over all answers, as in Eq. (\ref{eq:loss-nll}).
\begin{equation}\label{eq:loss-nll}
\mathcal{L}_{NLL} = -\frac{1}{m}\sum_{j=1}^m\sum_{t=1}^{L_\mathbf{y}}\log(p(y_t^{(j)}|y_{< t}^{(j)}, \mathbf{x}^{(j)}))
\end{equation}

\noindent
where $y_t^{(j)}$ is the $t$-th word in the $j$-th answer and $L_\mathbf{y}$ is the length of the gold answer. The loss inherently assumes that all answers in the dataset and dataset
are of the same quality, which is often not the case as shown by the example in Figure~\ref{f:zhidao}.

\section{Multi-Instance KBQA}

\subsection{Question Bags}
% One important challenge facing KBQA is the scarcity of high quality annotated data.
% Manually creating question-answer (QA) pairs according to a knowledge
% base requires a lot of effort~\cite{eric17kvincar}, while user-generated QA pairs can often
% be verbose, inaccurate, or irrelevant, although they are abundant online.
% Low quality answers can potentially misinform and hurt the model during training.
We start to depart from prior end-to-end KBQA efforts as we organize question-answer pairs
into question bags. A bag consists of a question and every answer to that question
in the dataset. We perform instance selection or weighting on the bag level, following the principle of multi-instance learning.

Consider the question bag in Figure~\ref{f:sys}.
There are four user-generated answers, A1--A4, towards the question ``What is the nickname of Cao Cao''? A3 and A4 try to answer the question directly with relevant knowledge. A2
covers two possible values (A-man, Mengde) of the KB
predicate. A1 is an uninformative answer. By organizing
Q and A1-A4 into a bag, we select or weight the answer instances according to their relevance
to the question, so that the model learns from A2-A4 but ideally not A1.

We define a QA bag to be a tuple $\mathcal{B}: <\mathbf{x},
\{\mathbf{y}^i\}>$, where $\{\mathbf{y}^i\}$ are the set of answers corresponding
to the question $\mathbf{x}$, and propose new loss functions to select or weight
the answers within a bag.
% The information in the bag is utilized in three different ways. \textit{Instance
% selection} dynamically selects one instance from each bag to train on; \textit{instance weighting}
% weights the instances in a bag according to their relative importance; \textit{curriculum
% learning} schedules the training process based on the characteristics of the bags.
\subsection{Answer Selection}

Similar to distantly supervised relation extractors, we first make the assumption that
at least one answer in the question bag is reasonable. Accordingly, instead of summing the loss
over all answers, we only train on one answer per bag which is the easiest to learn.
As is shown in Figure~\ref{f:sys}(a), only the loss on one of the answers will be
back-propagated, so uninformative answers like ``Check the history book yourself''
will not affect training, as they do not utilize KB information and are harder to generate.

Given QA bags $\mathcal{B}_1, \cdots, \mathcal{B}_n$, we define the minimum bag loss in
in Eq. (\ref{eq:loss-min-bag}).
\begin{equation}\label{eq:loss-min-bag}
\mathcal{L}_{SEL} = -\frac{1}{n}\sum_{k=1}^n\sum_{t=1}^{L_\mathbf{y}}\log(p(y_t^{*(k)}|y_{< t}^{*(k)}, \mathbf{x}^{(k)}))
\end{equation}

For each bag, we only calculate loss on one answer $y^*$ that is the closest to the network
output, as in Eq. (\ref{eq:loss-min-norm-bag-argmin}).
\begin{equation}\label{eq:loss-min-norm-bag-argmin}
y^{*(k)} = \argmax_{i, \mathbf{y}^i \in \mathcal{B}_k} LN_i\sum_{t=1}^{L_\mathbf{y}}\log(p(y_t^i|y_{< t}^i, \mathbf{x}^{(k)}))
\end{equation}

\noindent
note that we add a length normalizatoin term, $LN_i$, so that we do not
unjustly penalize long answers. Inspired by~\cite{wu2016google},
we set $LN_i=\frac{(5+1)^\alpha}{(5+L_\mathbf{y})^\alpha}$ ($\alpha=0.6$ in our
experiments).
% \subsubsection{Minimum Bag Loss with Length Normalization}
% % The minimum
% % bag level loss prefers short and simple answers, which is not always
% % the desired behavior for answer generation. For example, the highlighted
% % sentence ``Cao Mengde and Cao A-man'' in Figure~\ref{f:sys}(a) is unlikely
% % to be chosen due to
% % its length, although it is an informative answer.
% Minimum bag loss is To encourage the model to choose long and natural
% answers, we normalize answer lengths when choosing $y^*$.
% Formally, we calculate the loss according to Eq. (\ref{eq:loss-min-bag}),
% but choose $y^{*(k)}$ according to Eq. (\ref{eq:loss-min-norm-bag-argmin}).
% \noindent
% where $LN_i$ is the length normalization factor. Inspired by~\cite{wu2016google},
% we set $LN_i=\frac{(5+1)^\alpha}{(5+L_\mathbf{y})^\alpha}$ ($\alpha=0.6$ in our
% experiments).

\subsection{Answer Weighting}

Bag level minimum loss considers one instance from each bag. We also attempt to weight instances in a bag according to their
relevance to the question, so that every answer can contribute to the training
process, as in Figure~\ref{f:sys}(b).

The weight of an instance in a bag is then calculated based on consensus
among the answers,
as in Eq. (\ref{eq:loss-weighted}).
\vspace{-1mm}
\begin{equation}\label{eq:loss-weighted}
\mathcal{L}_{WGT} = -\frac{1}{n}\sum_{k=1}^n \sum_{i}\frac{C_i^{(k)}}{Z^{(k)}} \sum_{t=1}^{L_\mathbf{y}}\log(p(y_t^i|y_{< t}^i, \mathbf{x}^{(k)}))
\end{equation}

\noindent
where $i$ is the instance index within bag $k$, i.e. $\mathbf{y}^i \in \mathcal{B}_k$. $C_i^{(k)}$ is the weight for the $i^{th}$ instance in bag $\mathcal{B}_k$ (explained in following paragraphs), and
is normalized by $Z^{(k)}=\sum_i{C_i^{(k)}}$ in this work.

\paragraph{Content weighting.} We weight
the answers by their similarity to other answers in the same bag, assuming that
an unusual answer to a question is likely to be an outlier. Specifically,
we train a two-class Chinese InferSent~\cite{conneau2017infersent} model that predicts if two answers come from the same bag and encode
each answer in a bag with InferSent. We calculate cosine similarity among the answers.
The weight of an answer is its similarity to its nearest neighbor.
Specifically, denoting the InferSent encoding of an answer $a$ as $\textnormal{IS}_a$, $C_i^{(k)} = \argmax_{j} \cos (\textnormal{IS}_i^{(k)}, \textnormal{IS}_j^{(k)})$ in Eq. (\ref{eq:loss-weighted}), where $i$ and $j$ are answers in bag $\mathcal{B}_k$
and $i \neq j$.

\paragraph{KB weighting.} Content weighting does not take KB information into
account. To its remedy, we weight an answer instance by the importance of KB entities
mentioned by the answer. We
first measure the importance of an entity by its frequency in a bag. Consider the example in Figure~\ref{f:sys}.
Denoting ``A-man'' as $e_1$ and ``Mengde'' as $e_2$, we have entity count $c_{e1}=3$ (as ``A-man''
appears in A2, A3, and A4) and $c_{e2}=1$ (as ``Mengde'' appears in A2).
We then score an answer by the
sum of entity weights that occur in the answer: i.e. $C_i^{(k)} = \sum_{e \in \textbf{y}^i} c_e$
in Eq. (\ref{eq:loss-weighted}), favoring answers that mention more important
KB entities.
% This scheme favors answers that mention more important KB entities.

% \subsubsection{Entity average weighting} We can also score an answer by the average
% of entity weights within the sentence. i.e. $C_i^{(k)} = \frac{1}{|e|}\sum_{e \in \mathbf{y}^i} c_e$
% in Eq. (\ref{eq:loss-weighted}), where $|e|$ is the number of KB entities
% in $\mathbf{y}^i$. This scheme favors answers that mention only
% the popular entities in the KB.

% We also experiment with parameterized versions of instance weighting,
% similar in spirit to the selective attention approaches in relation
% extraction~\cite{lin2016selective}. Initial experiments show no
% improvement by the parameterized version.

\subsection{Curriculum Learning}\label{s:cl}
% The multi-instance learning approach needs more than one answers in a bag to effectively
% select or weight the answers.
% In practice, a large number of less popular questions have
%  single answers and it is still desirable to utilize these single instance bags in training.
Questions in real world datasets can have either one or multiple answers.
We schedule training under the curriculum learning principle,
as illustrated in Figure~\ref{f:sys}(c). Specifically,
assuming that we have both single- and multi-instance bags, train for $N$ iterations,
and the current
iteration is $N_c$, we always use the single-instance bags, but sample from multi-instance
bags with probability $(\frac{N_c}{N})^2$ in each iteration, so that we warm up training
with single instance bags in early iterations. Note that this is different from ~\cite{liu2018curriculum}, as they schedule single answers based on perceived difficulty, but we schedule question bags based on the number of answers within, which naturally indicates bag ambiguity.

\section{Experiments}

\subsection{Data Collection}

We experiment with two datasets in this paper: \textsc{CQA} and \textsc{PQA}.
\textsc{CQA}~\cite{he2017coreqa} is an open domain community QA dataset on
encyclopedic knowledge that is used by a number of KBQA systems~\cite{he2017coreqa,liu2018curriculum}.

We create a new dataset (\textsc{PQA}) in this paper.
\textsc{PQA} is a combination of user generated QA pairs and merchant-created
product KBs with relatively stable schema. We believe that it reflects a lot of
real world KBQA use cases.

% The detailed train / valid / test splits are shown in Table \ref{t:multi-datasets}.

\subsubsection{CQA}

CQA is collected from an encyclopedic community QA forum~\cite{he2017coreqa}.
The QA pairs are first collected from the forum and the KB is constructed
automatically. The questions and answers are then grounded to the KB.

\subsubsection{PQA: Data collection and processing}

\paragraph{Collection} \textsc{PQA} is a community product QA dataset we
collect from an e-commerce website. We focus on the mobile phone product domain,
because products have relatively stable KB schema: most products share
a number of ``core'' predicates, such as \texttt{display\_size}, \texttt{cpu\_model},
and \texttt{internal\_storage}. This is typical for product QA applications, but
not the case in \textsc{CQA}.
% In comparison, KB schema and user questions for domains such as clothing
% are much more sporadic.

We pre-select a number of popular mobile phone products and collect both the product
KB and the community QA pairs regarding the products.

% KBQA is
% best suited to answer the factual questions; answers for opinion questions are
% subjective and a best machine-generated answer might be the summarization of user
% opinions, which is out of the scope of this paper.
\paragraph{Filtering and preprocessing}
Product questions
can either be about facts (\textit{How large is the screen of this phone?}) or
opinions (``\textit{Does the screen look better than that of an iPhone?}'').
This paper works on factual questions only, so we build a
TextCNN~\cite{kim2014textcnn} classifier to filter the data. We manually annotate 1,000 questions
to train a binary classifier that obtains $0.86$ F1 on factual question detection.

We preprocess the KB to remove unit words (e.g. ``5.99 inch'' $\rightarrow$ ``5.99'')
from predicate values. We then ground the KB to QA pairs by string matching.

\paragraph{Differences from CQA} \textsc{PQA} is different from CQA in four aspects:
1) smaller size, 2) stable KB
schema, 3) the existence of ``advertising'' answers submitted by the merchants which are often irrelevant to the question, and
4) the KB attached to PQA consists of manually curated product properties,
instead of automatically extracted triples.
These features are more realistic for the product QA use case. We evaluate our method on
both data sets to more comprehensively measure its performance.

\subsection{Experimental Settings}
% \subsubsection{Implementation}
Our baseline is a re-implementation of CoreQA \cite{he2017coreqa} based on the released code.
%Our implementation is based on Tensorflow 1.4.0.
%All experiments are run on a Tesla M40 GPU accelerator.
% \begin{table}
% \begin{center}
% \begin{tabular}{c|c}
% Parameter      & Value\\ \hline
% Embedding Size & 200   \\
% Max Bag Size  & 5   \\
% KB State Size & 200   \\
% State Size & 600   \\
% Optimizer  & Adam \\
% Learning Rate & 0.001, decay: 0.96   \\
% % Layers Num & 2   \\
% \end{tabular}
% \end{center}
% \caption{\label{t:params} Experimental parameters}
% \end{table}
% \subsubsection{Parameters}

We follow the parameter setting of~\cite{he2017coreqa}
for the most part: we use embedding size of 200,
KB state size of 200, RNN state size of 600, and optimize with Adam~\cite{kingma2014adam}.
Embeddings are initiated randomly.
% . We limit a bag to have at most 5
%answers. The models are optimized with Adam, with learning rate of 0.001 and
% decay factor 0.96.

% \subsubsection{Metrics}
Following~\cite{liu2018curriculum}, we report Accuracy, BLEU(-2)~\cite{papineni2002bleu}, and Rouge(-L)~\cite{lin2004rouge}
to evaluate the adequacy and fluency of our outputs.
% We define accuracy as the percentage of correctly answered questions, using the terminology of~\cite{yin2016genqa}
% and~\cite{liu2018curriculum}. We follow~\cite{liu2018curriculum} to use
% character level BLEU-2 in our results, but note that as higher n-grams
% are not included in the BLEU-2 calculation, it still reflects more output adequacy (how
% much information is kept) but less output fluency (how natural the output is).
% We also report character-level Rouge (Rouge-L) scores, in order to more directly measure
% the naturalness of the output.

\begin{table}
\begin{center}
\begin{tabular}{r|ccc}
         &  Acc &  BLEU & ROUGE \\ \hline
CoreQA & 50.58 & 23.78 & 37.35 \\ \hline
%\makecell{Seq2Seq \\w/ attention} & 86.6 & \bf 81.9 & 84.2 \\
\textsc{Selection}  & \bf 67.69 & 28.35 &  35.86 \\
\textsc{-LengthNrm}  & 52.30 & 23.58 &  33.16 \\\hline
\textsc{Weight:KB}  & 60.45 & \bf 32.49 & \bf  38.88 \\
\textsc{Weight:Con}  & 56.75 & 28.92 & 37.05 \\
% \textsc{Weight:IS}   & 60.45 & \bf 32.49 & \bf  38.88 \\
% \textsc{Weight:Sum+IS}   & 60.45 & \bf 32.49 & \bf  38.88 \\
% \textsc{WGT-Avg}  & 58.23 & 31.38 & 37.54 \\ % line 7
\end{tabular}
\end{center}
\caption{\label{t:cqa-main} KBQA results: Multi-instance bags on CQA.}
\end{table}
\subsection{Multi-Instance Experiments}

We first evaluate the proposed methods on the multi-instance portion of the
CQA encyclopedic QA dataset released by~\cite{he2017coreqa}, as our method
is designed to work with questions with multiple answers. The dataset consists of 64K bags for training/validation and 16K bags for testing. Each question bag has 3.2 answers on average.
As shown in Table~\ref{t:cqa-main},
both answer selection (\textsc{Selection}) and weighting (\textsc{Weight:KB}) obtain better
accuracy than the baseline (CoreQA). Instance selection
leads to 17.11 absolute point improvement in accuracy, but no improvement on Rouge.
We hypothesize that instance selection limits generation naturalness, because the models are in effect trained on less examples. Answer weighting achieves
better balance between adequacy and fluency, improving accuracy by 9.87 absolute points
and Rouge by 1.53 absolute points. We use KB weighting in following
experiments.

We also measured the performance of answer selection
without length normalization (\textsc{-LengthNrm}) and answer weighting using
content instead of KB information (\textsc{Weight:Con}) for ablation study.
The results
confirm the necessity of performing length normalization and utilizing KB information
in answer weighting.

\begin{table}
\begin{center}
\begin{tabular}{c|c|ccc}
& System & Acc & BLEU & ROUGE \\ \hline
%\makecell{Seq2Seq \\w/ attention} & 86.6 & \bf 81.9 & 84.2 \\
% \multirow{3}{*}{\rotatebox[origin=c]{90}{\makecell{\textsc{CQA} \\ \textsc{Mixed}}}}
% & CoreQA        & 56.89 & 25.92 & 37.66 \\
% & WGT-Multi     & 60.27 & 28.67 & 35.14 \\
% & WGT-Mixed     & \bf{64.60} & \bf{30.00} & \bf{37.76} \\
% & Curriculum    & 62.63 & 29.16 & 36.46 \\\hline
\multirow{3}{*}{\rotatebox[origin=c]{90}{\textsc{CQA}}}
& \cite{he2017coreqa} & 56.6 & -  & - \\
& CoreQA        & 57.58 & 27.76 & 37.67 \\
% & WGT-Multi     & 63.33 & 26.08 & 31.53 \\
& \textsc{Weight}     & 70.18 & \bf{28.88} & 36.78 \\
& Curriculum    & \bf{71.56} & 27.21 & \bf{39.25} \\\hline
\multirow{3}{*}{\rotatebox[origin=c]{90}{\textsc{PQA}}}
& CoreQA       & 52.50 & 18.63 & 32.55 \\
% & WGT-Multi    & 54.14 & 18.83 & 32.97  \\
& \textsc{Weight}    & 59.92 & 22.02 & 32.73  \\
& Curriculum   & \bf{64.27} & \bf{22.85} & \bf{33.47}  \\ % single->multi
% & Anti-curriculum   & 64.66  & 22.14 & 34.30  \\ % multi->single
\end{tabular}
\end{center}
\caption{\label{t:all-test-instances} KBQA results: Complete data on CQA/PQA.}
\end{table}

\subsection{Mixed-Instance Experiments}

We next experiment on complete real word datasets that contain both single- and
multiple-instance bags. In addition to CQA, we also evaluate on a product QA dataset
(PQA) that we collect from a large e-commerce website, with 87K question bags for
training/validation and 22K bags for testing. Half of the questions have multiple
answers, with 2.4 answers on average.
Unlike CQA, which automatically extracts the KB, the PQA KB is composed
of real product properties on the website and is closer to
the production KBQA setting.

In Table~\ref{t:all-test-instances}, the first line is the result reported by~\cite{he2017coreqa}. CoreQA is
our re-implementation of~\cite{he2017coreqa}, \textsc{Weight} is trained with KB-based
instance weighting and is identical to the \textsc{Weight:KB} setting in Table~\ref{t:cqa-main}, and Curriculum trains first on single instance bags and gradually adds
multi-instance bags (cf. \S~\ref{s:cl}).

On both datasets, instance weighting significantly improves the
accuracy scores and curriculum learning further improves both accuracy and
naturalness. On the CQA public dataset, instance weighting with
a curriculum schedule leads to 13.98 and 1.58 absolute points improvement on
accuracy and ROUGE respectively. The trend is similar for the PQA dataset,
showing that the method works for different domains and use cases.

\section{Conclusions and Future Work}

We trained end-to-end KBQA models with multi-instance learning principles. We showed that selection and weighting of answers to the same question helps
reducing noise in the training data and boosts both output adequacy and naturalness.
% We first organized QA pairs into question bags and then performed dynamic instance selection/weighting at the bag level. We also applied curriculum learning to schedule bags of different sizes.
% We performed extensive experiments on a public encyclopedic QA data set as
% well as an
% in-house product QA dataset and showed that multi-instance learning improved KBQA performance.
% When training on both questions with multiple answers and those with single answers,
% multi-instance learning still helped answer accuracy and naturalness.

Our approach is independent of the underlying QA model. In future,
we plan to integrate our approach with more QA models and explore more
ways to utilize the information in the bag.

\bibliography{main}
\bibliographystyle{acl}

\end{document}